\documentclass[conference]{IEEEtran}
\IEEEoverridecommandlockouts
\usepackage{cite}
\usepackage{amsmath,amssymb,amsfonts}
\usepackage{algorithmic}
\usepackage{graphicx}
\usepackage{textcomp}
\usepackage{xcolor}
\usepackage{multirow}
\usepackage{multicol}
\usepackage{array}
\usepackage{graphicx}
\usepackage{subcaption}
\usepackage{stfloats}
\usepackage[hidelinks]{hyperref}

\definecolor{repoBlue}{RGB}{0,102,204}

\def\BibTeX{{\rm B\kern-.05em{\sc i\kern-.025em b}\kern-.08em
    T\kern-.1667em\lower.7ex\hbox{E}\kern-.125emX}}
\begin{document}

\title{Benchmarking UAV-based Vehicle Re-Identification under Simulated Weather Conditions}


\author{
\IEEEauthorblockN{1\textsuperscript{st} Vu Minh Tran
\IEEEauthorblockA{
\textit{University of Information Technology} \\
\textit{Vietnam National University Ho Chi Minh City} \\
Ho Chi Minh City, Vietnam \\
23521819@gm.uit.edu.vn
}
}

\and
\IEEEauthorblockN{2\textsuperscript{nd} Khang Nguyen\textsuperscript{*}}
\IEEEauthorblockA{
\textit{University of Information Technology} \\
\textit{Vietnam National University Ho Chi Minh City} \\
Ho Chi Minh City, Vietnam \\
khangnttm@uit.edu.vn
}
\thanks{\textsuperscript{*} Corresponding author: Khang Nguyen, khangnttm@uit.edu.vn.}
}

\maketitle

\begin{abstract}
UAV-based vehicle re-identification (ReID) has emerged as a promising technique for traffic surveillance, urban monitoring, and public-safety applications thanks to the flexible viewpoints and wide-area coverage provided by unmanned aerial vehicles. However, despite recent progress on UAV-based vehicle ReID benchmarks, the robustness of existing methods under adverse weather remains insufficiently studied. This is important because weather degradation can significantly affect the fine-grained appearance cues required for reliable vehicle matching in aerial imagery, especially under small object scale, viewpoint variation, and complex backgrounds. In this paper, we present a controlled comparative study of three representative recent vehicle ReID methods, namely CLIP-ReID, MSINet, and AdaSP, on two UAV-based benchmarks, VRU and UAV-VeID. To ensure consistent robustness evaluation, we generate synthetic foggy and rainy variants of both datasets using an analytical weather-effect pipeline while preserving the original identities and data splits. All methods are then trained and evaluated under matched clean, foggy, and rainy conditions. Experimental results show that adverse weather consistently degrades retrieval performance across both datasets, with rain causing larger drops than fog in nearly all settings. Among the evaluated methods, AdaSP demonstrates the strongest robustness, achieving 93.0\% and 88.5\% mAP on VRU-Large, and 88.7\% and 76.2\% mAP on UAV-VeID-Test under foggy and rainy conditions, respectively. Overall, our findings show that simulated adverse weather substantially increases the difficulty of UAV-based vehicle ReID, reveals clear robustness differences among recent methods, and highlights the need for weather-aware model design and evaluation protocols in future aerial ReID research. The code is released at \href{https://github.com/tranminhvu945/Benchmarking-ReID}{\textcolor{repoBlue}{https://github.com/tranminhvu945/Benchmarking-ReID}}.
\end{abstract}


\section{Introduction}
\begin{figure*}[!ht]
    \centering
    \includegraphics[scale=0.16]{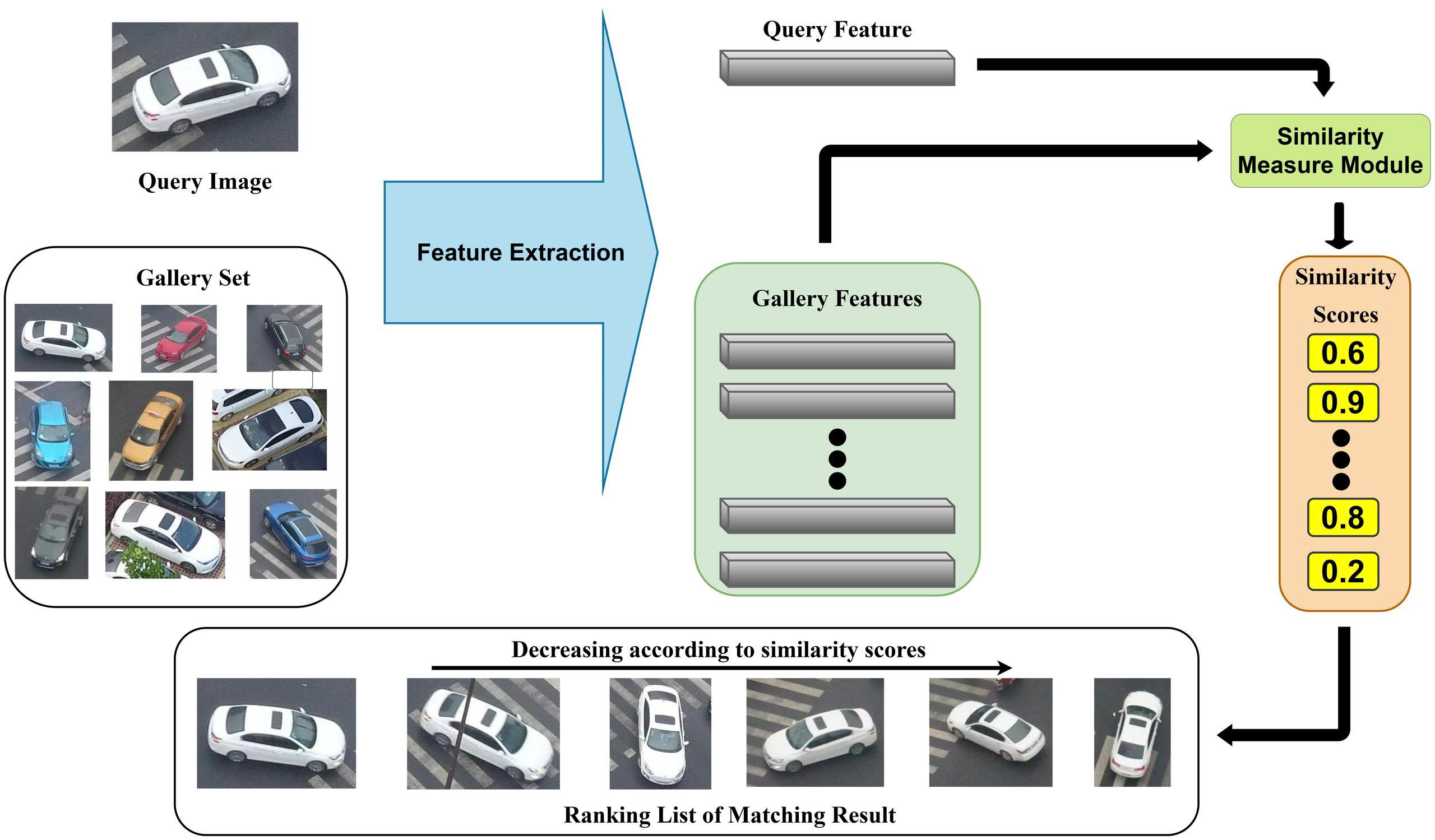}
    \caption{Illustration of the vehicle re-identification pipeline. A query image is matched against a gallery set by extracting visual features and ranking gallery samples according to similarity scores.}
    \label{fig:reid_pipeline}
\end{figure*}

Vehicle re-identification (ReID) aims to match the same vehicle across different images or camera views and has become an important research topic in computer vision due to its applications in intelligent transportation, large-scale surveillance, traffic analysis, and public safety \cite{wang2024survey}. As illustrated in Fig.~\ref{fig:reid_pipeline}, a typical vehicle ReID system takes a query image, compares it against a gallery set in a learned feature space, and returns a ranked list of candidate matches according to similarity scores. In recent years, unmanned aerial vehicles (UAVs) have expanded the scope of this problem by enabling image acquisition from flexible viewpoints, varying altitudes, and dynamic trajectories. Compared with conventional fixed-camera settings, UAV-based vehicle ReID is inherently more challenging because the same vehicle may appear under large viewpoint shifts, scale variation, background clutter, motion blur, and unstable observation conditions \cite{lu2022vru,teng2021uavveid,wang2019vrai}. These challenges are also consistent with observations reported in recent UAV-oriented re-identification surveys \cite{albaluchi2024uavreid}.

Despite substantial progress in vehicle ReID under clean conditions, practical UAV deployment must also consider environmental corruption. Among these factors, adverse weather is particularly important. Fog reduces visibility and contrast, while rain introduces streaks, blur, and appearance distortion. These effects weaken the discriminative visual cues needed for fine-grained matching, such as roof contour, windshield shape, window geometry, and local texture. In UAV imagery, the problem is even more pronounced because aerial images already contain smaller object scales and more severe perspective variation than ground-based surveillance images. Analytical weather simulation has recently become a practical tool for studying robustness under controlled corruption while preserving scene semantics \cite{gupta2024weather}. 

Recent representative methods such as CLIP-ReID \cite{li2023clipreid}, MSINet \cite{gu2023msinet}, and AdaSP \cite{zhou2023adasp} have shown strong performance in image or object re-identification. In particular, CLIP-ReID builds on the transferable visual representations introduced by CLIP \cite{radford2021clip}. However, these methods are typically analyzed from the perspective of architecture design, metric learning, or feature representation, whereas a focused study of their robustness under controlled weather degradation remains limited for UAV vehicle ReID. This gap makes it difficult to understand how existing methods behave when image quality deteriorates in aerial scenarios that are closer to real deployment.

To address this issue, this paper revisits UAV vehicle ReID from a weather-aware comparative perspective. Rather than proposing a new architecture, we conduct a controlled evaluation of CLIP-ReID, MSINet, and AdaSP on the VRU and UAV-VeID datasets under normal, foggy, and rainy conditions. Foggy and rainy images are generated using an analytical weather-effect generation method \cite{gupta2024weather}, allowing the original identities and dataset splits to be preserved across weather conditions. For each condition, training and evaluation are performed on the corresponding weather-specific split. This setting allows us to compare how representative ReID methods perform under matched clean, foggy, and rainy conditions and to analyze how simulated weather affects retrieval difficulty in UAV imagery.

Our main contributions are as follows:
\begin{itemize}
    \item We establish a controlled evaluation setting for UAV vehicle ReID under synthetic fog and rain using VRU and UAV-VeID.
    \item We compare three representative recent methods, namely CLIP-ReID, MSINet, and AdaSP, under matched clean, foggy, and rainy training/evaluation settings using the same datasets, metrics, and weather-generation pipeline.
    \item We report quantitative degradation trends, drop-pattern analysis, and representative retrieval failures, showing that adverse weather remains a major challenge for UAV-based vehicle retrieval.
\end{itemize}

\section{Related Work}
Vehicle ReID has been widely studied in fixed-camera surveillance, where CNN-based, transformer-based, and metric-learning methods have achieved strong performance under relatively clean conditions~\cite{wang2024survey}. However, much of this literature focuses on ground-based surveillance settings, where viewpoint ranges are narrower and benchmark images are typically cleaner than those captured by UAV platforms.

UAV-based vehicle ReID introduces a more difficult setting. Compared with ground cameras, UAVs capture vehicles from oblique and top-down views, over wider height ranges and with more unstable backgrounds. Datasets such as UAV-VeID \cite{teng2021uavveid}, VRU \cite{lu2022vru}, and VRAI \cite{wang2019vrai} have enabled the study of re-identification in aerial imagery, where the appearance of the same vehicle may change significantly across flight altitude, camera angle, and scene context. These datasets better reflect real-world aerial monitoring scenarios, but robustness under weather corruption remains insufficiently characterized.

Among recent representative methods, CLIP-ReID \cite{li2023clipreid} leverages CLIP-based visual representations \cite{radford2021clip} to improve image re-identification without requiring concrete text labels. MSINet \cite{gu2023msinet} integrates multi-scale interaction with architecture search to construct an efficient object ReID backbone. AdaSP \cite{zhou2023adasp} improves pairwise metric learning by adaptively balancing hardest and less-hard positive pairs during training. These methods represent three complementary design directions: pretrained visual representation learning, searched multi-scale interaction, and adaptive metric learning. This makes them suitable choices for a robustness-oriented comparative study.

Weather corruption has recently received increasing attention in general vision tasks, including object detection and recognition. Analytical simulation methods provide a practical way to evaluate robustness under controlled corruption while preserving the underlying scene semantics \cite{gupta2024weather}. Beyond adverse-weather synthesis, robustness under common image corruptions has also been studied as an important evaluation dimension for modern visual recognition systems \cite{hendrycks2019benchmarking}.

Inspired by this direction, our work studies how recent UAV vehicle ReID methods behave when the same identities are observed under simulated fog and rain. In this sense, the paper contributes not a new model, but a focused robustness analysis for aerial vehicle retrieval under adverse weather.

\section{Compared Methods and Evaluation Protocol}
\subsection{Compared Methods}
We evaluate three representative recent methods: CLIP-ReID, MSINet, and AdaSP.

\begin{itemize}
    \item \textbf{CLIP-ReID:} \cite{li2023clipreid} is a vision-language inspired ReID framework built on CLIP representations \cite{radford2021clip}. In this study, we report its two commonly used image encoders, namely ResNet-50 and ViT-B/16. CLIP-ReID is particularly relevant to this study because it represents a pretrained visual representation learning paradigm, and its behavior under aerial weather corruption offers useful insight into the robustness of large-scale pretrained features.
    \item \textbf{MSINet:} \cite{gu2023msinet} is a multi-scale interaction network designed for object ReID through architecture search and contrastive optimization. Its design explicitly captures multi-scale information and cross-branch interaction, which may help maintain discriminative representations when local details are partially degraded by fog or rain.
    \item \textbf{AdaSP:} \cite{zhou2023adasp} is a metric-learning approach based on adaptive sparse pairwise loss. By balancing hardest and less-hard positive pairs during training, AdaSP encourages more stable feature discrimination and is therefore a strong candidate for robustness under appearance corruption.
\end{itemize}
These methods were selected because they reflect complementary design directions: pretrained visual representation learning, searched multi-scale interaction, and adaptive metric learning.

\subsection{Datasets and Weather Generation}
\begin{table}[!h]
\centering
\caption{Summary of datasets and evaluation splits}
\label{tab:dataset_summary}
\renewcommand{\arraystretch}{1.3}
\resizebox{\columnwidth}{!}{
\begin{tabular}{|c|c|c|c|c|c|c|}
    \hline
    \textbf{Dataset} & \textbf{Split} & \textbf{Images} & \textbf{Identities} & \textbf{Surveillance} & \textbf{Multi-view} & \textbf{Multi-scale} \\ \hline
    \multirow{4}{*}{VRU} & Train & 80,532 & 7,085 & \multirow{4}{*}{5 UAVs} & \multirow{4}{*}{$\checkmark$} & \multirow{4}{*}{$\checkmark$} \\ \cline{2-4}
     & Small & 13,920 & 1,200 & & & \\ \cline{2-4}
     & Medium & 27,345 & 2,400 & & & \\ \cline{2-4}
     & Large & 91,595 & 8,000 & & & \\ \hline
    \multirow{3}{*}{UAV-VeID} & Train & 18,709 & 1,797 & \multirow{3}{*}{2 UAVs} & \multirow{3}{*}{$\checkmark$} & \multirow{3}{*}{$\checkmark$} \\ \cline{2-4}
     & Validation & 4,150 & 596 & & & \\ \cline{2-4}
     & Test & 19,058 & 2,208 & & & \\ \hline
    \end{tabular}
}
\end{table}

\textbf{Datasets:} We use two UAV-based vehicle ReID datasets in this study: VRU \cite{lu2022vru} and UAV-VeID \cite{teng2021uavveid}. VRU is a large-scale UAV vehicle ReID dataset containing substantial variation in viewpoint, scale, and scene context. UAV-VeID is another aerial dataset designed for vehicle re-identification under realistic UAV capture conditions. Detailed statistics of the two datasets, including the evaluation splits and the numbers of images and identities, are summarized in Table~\ref{tab:dataset_summary}.

\begin{figure*}[!t]
\centering
\includegraphics[width=0.75\textwidth]{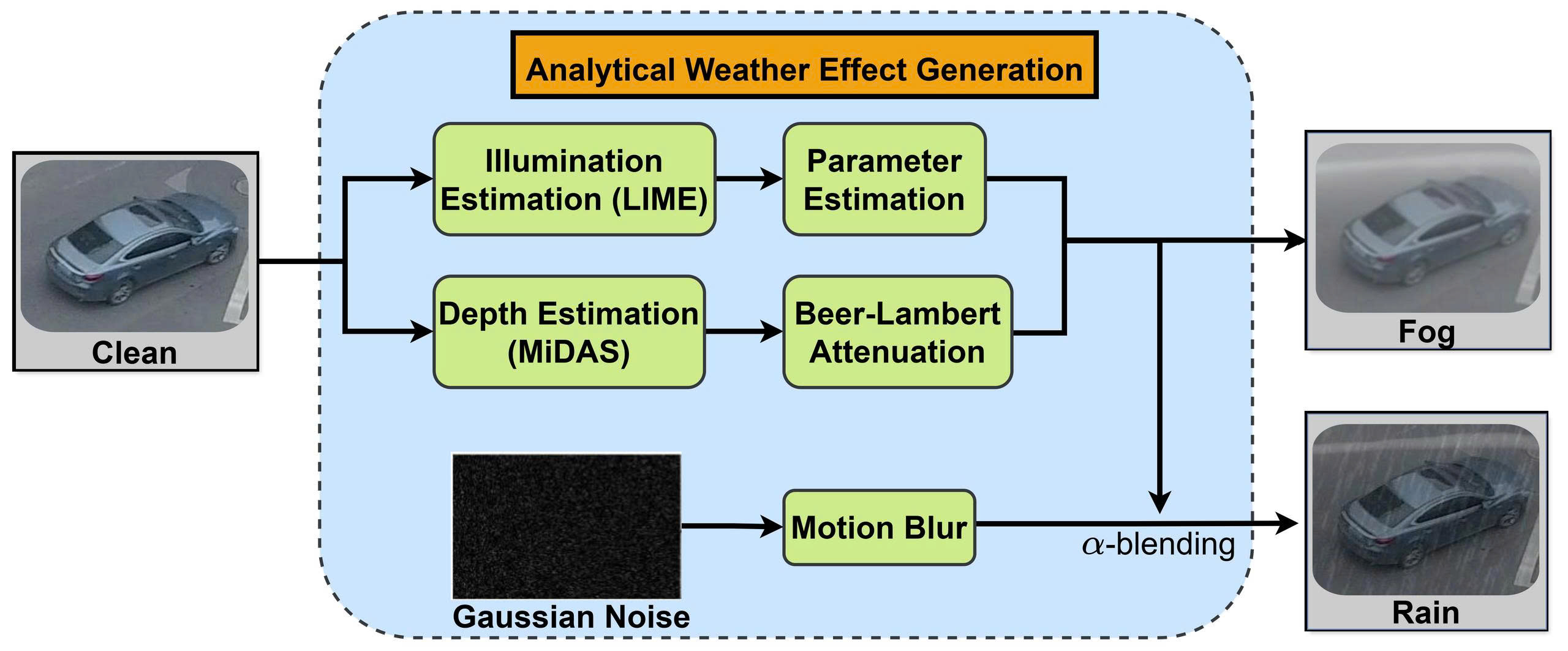}
\caption{Overview of the analytical weather-effect generation pipeline. }
\label{fig:gen}
\end{figure*}

\textbf{Weather Generation:} To study the effect of weather conditions in a controlled manner, we generate foggy and rainy variants of both datasets using the analytical weather-effect generation method in \cite{gupta2024weather}. Fig.~\ref{fig:gen} illustrates the analytical weather-effect generation process used in this study. Starting from a clean UAV image, the method first estimates scene cues through illumination estimation (LIME) and depth estimation (MiDaS). The estimated illumination is used to derive weather-related parameters, while the depth map is used in the Beer--Lambert attenuation step to synthesize fog. For rain generation, Gaussian noise is transformed into rain streaks through motion blur, and the resulting streak pattern is combined with the weather-degraded base image by alpha blending to produce the final rainy image.

For each clean image, we generate one foggy version and one rainy version using a fixed weather-generation configuration for the corresponding condition. The same generation protocol is applied consistently across the train, validation, and test splits of each dataset to avoid condition-specific bias. Consequently, each dataset is available in three matched versions: clean, foggy, and rainy.

In our experiments, models are trained and evaluated within the same weather condition, enabling a controlled comparison under matched weather-specific settings.

\subsection{Evaluation Metrics \& Implementation Details}
\textbf{Metrics:} In this paper, we utilize two commonly used metrics in the Re-ID domain: mAP (mean Average Precision) and CMC@K (Cumulative Matching Characteristic).

\begin{itemize}
    \item \textbf{mAP:} This metric reflects how well the model ranks all relevant matches across the entire retrieval list. For a query $q$, let $N_q$ denote the number of relevant gallery images, $\mathrm{rel}_q(k) \in \{0,1\}$ indicate whether the item ranked at position $k$ is relevant, and $P_q(k)$ denote the precision computed up to rank $k$. The Average Precision (AP) for query $q$ is defined as
    \begin{equation}
    AP(q)=\frac{1}{N_q}\sum_{k=1}^{n} P_q(k)\,\mathrm{rel}_q(k),
    \end{equation}
    where $n$ is the size of the gallery. The mean Average Precision over all $Q$ queries is then computed as
    \begin{equation}
    mAP=\frac{1}{Q}\sum_{q=1}^{Q} AP(q).
    \end{equation}

    \item \textbf{CMC@K:} T his metric  measures whether at least one correct match appears within the top-$K$ retrieved results. Let $r_q$ denote the rank position of the first correct match for query $q$. Then CMC at rank $K$ is defined as
    \begin{equation}
    CMC@K=\frac{1}{Q}\sum_{q=1}^{Q}\mathbf{1}(r_q \leq K),
    \end{equation}
    where $\mathbf{1}(\cdot)$ is the indicator function. In this study, we use Rank 1 and Rank 5, corresponding to K=1 and K=5, respectively.

\end{itemize}

\textbf{Implementation:} We use the official implementations of CLIP-ReID, MSINet, and AdaSP and follow the default training settings reported in their original papers for all three methods \cite{li2023clipreid,gu2023msinet,zhou2023adasp}. No method-specific hyperparameter retuning is introduced in order to maintain a fair and reproducible comparison. Experiments are conducted on a server equipped with NVIDIA GeForce RTX 2080 Ti GPUs.

\section{Results and Discussion}

\begin{table*}[!b]
\centering
\caption{Performance comparison on the VRU dataset. \textcolor{red}{Red} and \textcolor{blue}{blue} indicate the best and second-best results, respectively.}
\label{tab:vru_results}
\renewcommand{\arraystretch}{1.3}
\resizebox{\textwidth}{!}{
\begin{tabular}{|c|c|cc|cc|cc|cc|cc|cc|cc|cc|cc|}
\hline
\multirow{3}{*}{\textbf{Method}} & \multirow{3}{*}{\textbf{Backbone}} & \multicolumn{6}{c|}{\textbf{Normal}} & \multicolumn{6}{c|}{\textbf{Foggy}} & \multicolumn{6}{c|}{\textbf{Rainy}} \\ \cline{3-20} 
 &  & \multicolumn{2}{c|}{\textbf{Small}} & \multicolumn{2}{c|}{\textbf{Medium}} & \multicolumn{2}{c|}{\textbf{Large}} & \multicolumn{2}{c|}{\textbf{Small}} & \multicolumn{2}{c|}{\textbf{Medium}} & \multicolumn{2}{c|}{\textbf{Large}} & \multicolumn{2}{c|}{\textbf{Small}} & \multicolumn{2}{c|}{\textbf{Medium}} & \multicolumn{2}{c|}{\textbf{Large}} \\ \cline{3-20} 
 &  & mAP & R1 & mAP & R1 & mAP & R1 & mAP & R1 & mAP & R1 & mAP & R1 & mAP & R1 & mAP & R1 & mAP & R1 \\ \hline
\multirow{2}{*}{\textbf{CLIP-ReID}} & ResNet-50 & \color{blue}{98.8} & 97.9 & \color{blue}{97.9} & 96.6 & \color{blue}{95.2} & 92.3 & \color{blue}{97.4} & 95.6 & \color{blue}{95.5} & 92.7 & \color{blue}{91.2} & 86.5 & \color{blue}{94.5} & 91.5 & \color{blue}{92.1} & 88.5 & \color{blue}{86.8} & 81.6 \\ \cline{2-20} 
 & ViT-B/16 & 97.7 & 95.9 & 95.8 & 93.0 & 91.4 & 86.3 & 93.9 & 89.5 & 90.1 & 84.2 & 81.7 & 72.7 & 91.1 & 86.6 & 87.0 & 80.9 & 78.7 & 70.4 \\ \hline
\textbf{MSINet} & - & 98.2 & 96.9 & 97.0 & 94.9 & 93.8 & 90.0 & 96.9 & 94.7 & 94.8 & 91.5 & 89.9 & 84.5 & 93.1 & 89.7 & 90.2 & 85.6 & 84.1 & 77.9 \\ \hline
\textbf{AdaSP} & ResNet-50 + IBN & \color{red}{99.1} & 98.4 & \color{red}{98.2} & 97.0 & \color{red}{95.9} & 93.3 & \color{red}{97.9} & 96.4 & \color{red}{96.6} & 94.4 & \color{red}{93.0} & 89.2 & \color{red}{95.2} & 92.7 & \color{red}{93.4} & 90.1 & \color{red}{88.5} & 83.7 \\ \hline
\end{tabular}
}
\end{table*}

\begin{table*}[!b]
\centering
\caption{Performance comparison on the UAV-VeID dataset. \textcolor{red}{Red} and \textcolor{blue}{blue} indicate the best and second-best results, respectively.}
\label{tab:uav-veid_results}
\renewcommand{\arraystretch}{1.3} 
\resizebox{\textwidth}{!}{
\begin{tabular}{|c|c|ccc|ccc|ccc|ccc|ccc|ccc|}
\hline
 & & \multicolumn{6}{c|}{\textbf{Normal}} & \multicolumn{6}{c|}{\textbf{Foggy}} & \multicolumn{6}{c|}{\textbf{Rain}} \\ \cline{3-20} 
 & & \multicolumn{3}{c|}{\textbf{Validation}} & \multicolumn{3}{c|}{\textbf{Test}} & \multicolumn{3}{c|}{\textbf{Validation}} & \multicolumn{3}{c|}{\textbf{Test}} & \multicolumn{3}{c|}{\textbf{Validation}} & \multicolumn{3}{c|}{\textbf{Test}} \\ \cline{3-20} 
\multirow{-3}{*}{\textbf{Method}} & \multirow{-3}{*}{\textbf{Backbone}} & mAP & R1 & R5 & mAP & R1 & R5 & mAP & R1 & R5 & mAP & R1 & R5 & mAP & R1 & R5 & mAP & R1 & R5 \\ \hline
 & ResNet-50 & {\color{blue} 96.4} & 94.2 & 99.0 & {\color{blue} 88.1} & 83.7 & 93.4 & {\color{blue} 95.6} & 92.8 & 98.9 & {\color{blue} 83.5} & 77.7 & 90.4 & {\color{blue} 89.1} & 85.3 & 93.7 & {\color{blue} 72.4} & 66.9 & 78.4 \\ \cline{2-20} 
\multirow{-2}{*}{\textbf{CLIP-ReID}} & ViT-B/16 & 91.2 & 86.6 & 96.9 & 80.1 & 73.0 & 88.7 & 84.1 & 76.0 & 94.5 & 70.4 & 61.4 & 81.1 & 79.2 & 71.5 & 88.5 & 63.3 & 56.3 & 71.1 \\ \hline
\textbf{MSINet} & - & 96.2 & 94.3 & 98.6 & 87.7 & 82.6 & 94.1 & 94.5 & 91.8 & 98.1 & 82.5 & 76.0 & 90.3 & 88.5 & 84.5 & 93.2 & 71.0 & 64.8 & 78.0 \\ \hline
\textbf{AdaSP} & ResNet-50 + IBN & {\color{red} 97.1} & 95.4 & 99.1 & {\color{red} 92.7} & 89.4 & 96.7 & {\color{red} 96.6} & 94.6 & 99.0 & {\color{red} 88.7} & 84.1 & 94.3 & {\color{red} 91.2} & 88.2 & 94.8 & {\color{red} 76.2} & 70.9 & 82.1 \\ \hline
\end{tabular}
}
\end{table*}

\subsection{Main Results under Normal Conditions}
Tables \ref{tab:vru_results} and \ref{tab:uav-veid_results} summarize the quantitative results on VRU and UAV-VeID under normal, foggy, and rainy conditions. Under normal conditions, all methods achieve strong retrieval performance, confirming that current ReID approaches can learn highly discriminative vehicle representations from UAV imagery when weather corruption is absent. However, the relative differences between methods become more informative on the more challenging evaluation splits.

On VRU, AdaSP achieves the best normal-condition performance across all three test splits, reaching 99.1\%, 98.2\%, and 95.9\% mAP on Small, Medium, and Large, respectively. On the same dataset, CLIP-ReID with ResNet-50 remains highly competitive, obtaining 98.8\%, 97.9\%, and 95.2\% mAP, while MSINet reaches 98.2\%, 97.0\%, and 93.8\%. CLIP-ReID with ViT-B/16 consistently trails the other variants, especially on VRU-Large, where it records 91.4\% mAP and 86.3\% Rank-1.

A similar trend can be observed on UAV-VeID. AdaSP again delivers the strongest performance, achieving 97.1\% mAP on Validation and 92.7\% mAP on Test. CLIP-ReID with ResNet-50 achieves 96.4\% and 88.1\% mAP on the two splits, while MSINet obtains 96.2\% and 87.7\%. CLIP-ReID with ViT-B/16 performs noticeably worse, especially on the Test split, where it reaches 80.1\% mAP and 73.0\% Rank-1.

Two observations are worth noting. First, performance on easier splits such as VRU-Small and UAV-VeID-Validation is close to saturation for several methods, which compresses the visible gap between them. Second, the gap widens on harder splits such as VRU-Large and UAV-VeID-Test, suggesting that these settings are more suitable for revealing differences in feature robustness and generalization. In addition, the ResNet-50 variant of CLIP-ReID is consistently more stable than the ViT-B/16 variant in our setting, indicating that under the current UAV data scale and training protocol, the convolutional backbone provides a more reliable trade-off between representation strength and robustness.

\begin{figure*}[b!]
    \centering
    \begin{minipage}{0.48\textwidth}
        \centering
        \includegraphics[width=\linewidth]{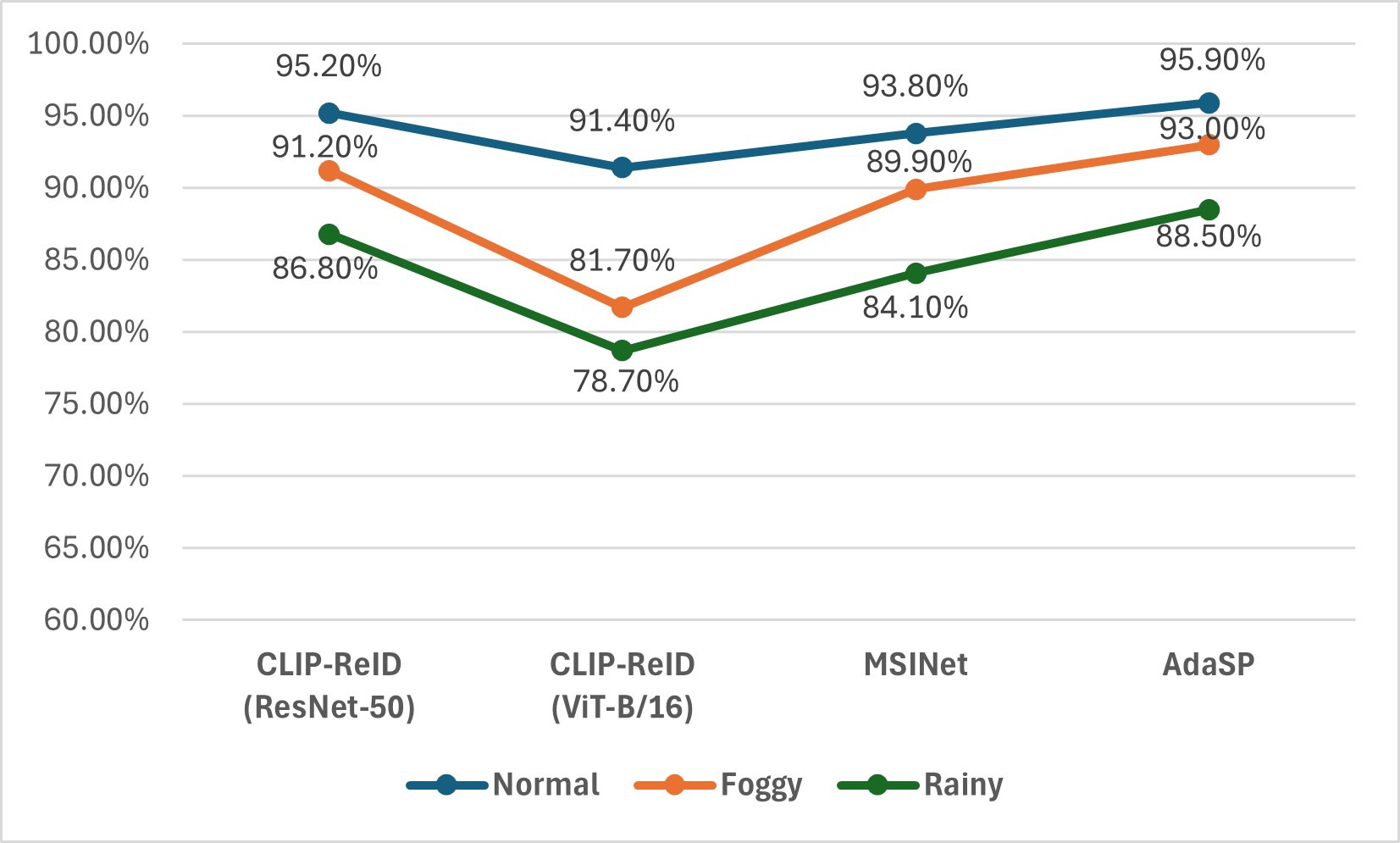}
        \subcaption{Results on VRU Dataset (Large)}
        \label{fig:VRU_Big}
    \end{minipage}
    \begin{minipage}{0.48\textwidth}
        \centering
        \includegraphics[width=\linewidth]{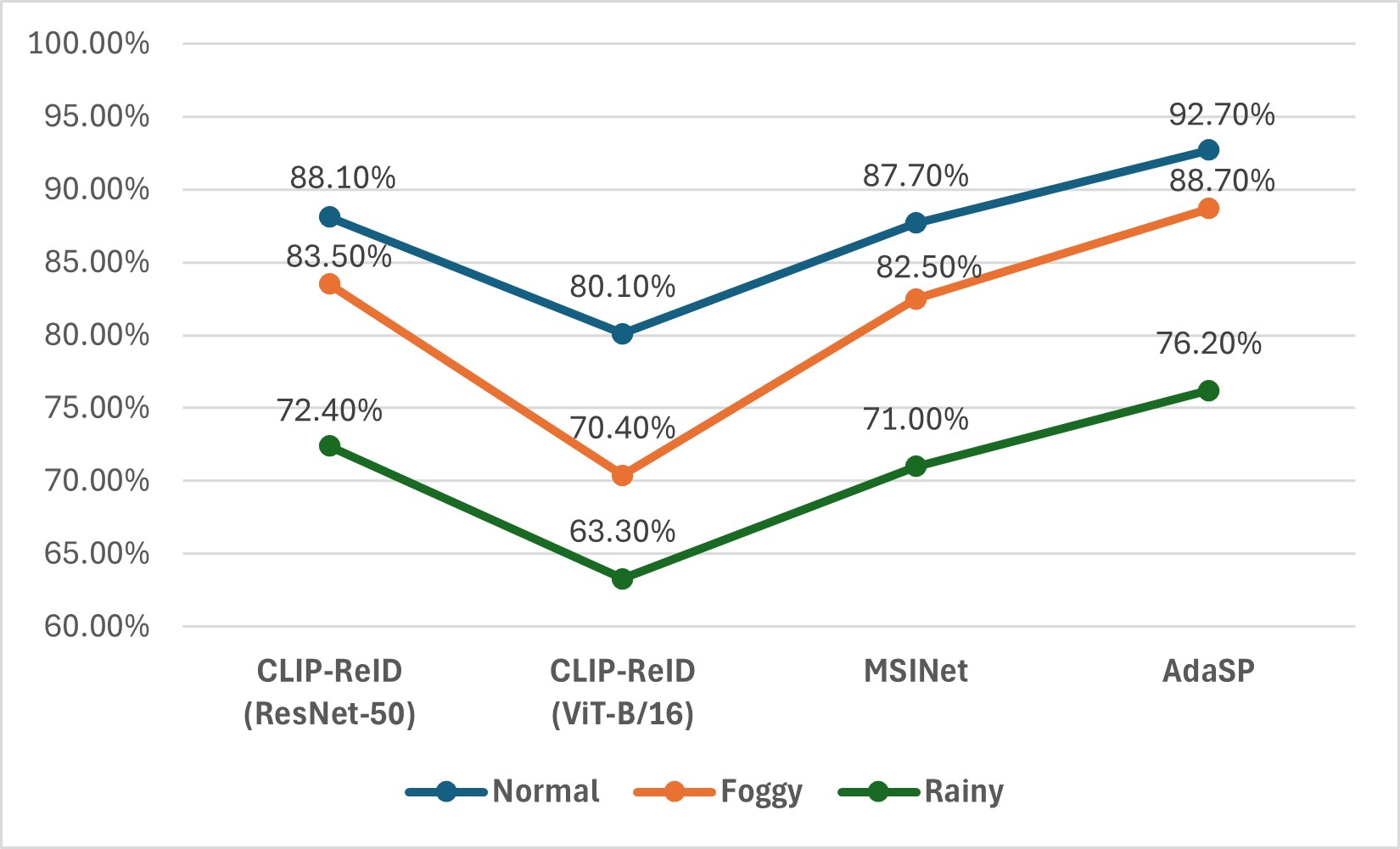}
        \subcaption{Result on UAV-VeID Dataset (Test)}
        \label{fig:UAV_VeID_Test}
    \end{minipage}
    \caption{mAP trends on the two most challenging evaluation settings under normal, foggy, and rainy conditions.}
    \label{fig:chart}
\end{figure*}

\begin{figure*}[b!]
    \centering
    \includegraphics[width=\linewidth]{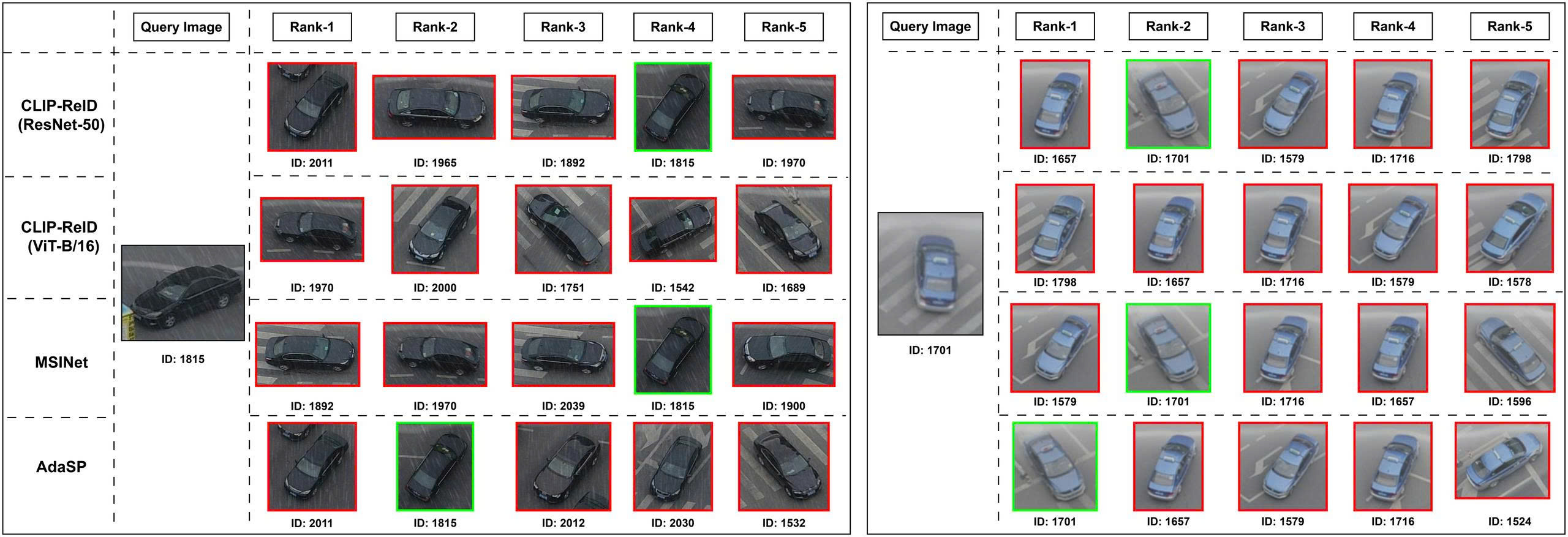}
    \caption{Qualitative Rank-5 retrieval results under adverse weather conditions. \textcolor{green}{Green} and \textcolor{red}{red} borders denote correct and incorrect matches, respectively.}
    \label{fig:Exaple_result}
\end{figure*}

\subsection{Performance under Foggy and Rainy Conditions}
When foggy and rainy corruption is introduced, all methods exhibit a clear performance decline. However, the magnitude of degradation differs substantially across methods and across datasets. In general, fog causes a moderate reduction, whereas rain leads to a much larger performance drop on both VRU-Large and UAV-VeID-Test. This trend suggests that rain more severely disrupts the fine-grained visual cues required for vehicle matching in aerial imagery.

On VRU-Large, AdaSP shows the strongest robustness. Its mAP decreases from 95.9\% under normal conditions to 93.0\% under fog and 88.5\% under rain, corresponding to absolute drops of 2.9 and 7.4 points, respectively. In comparison, CLIP-ReID with ResNet-50 drops from 95.2\% to 91.2\% and 86.8\%, MSINet drops from 93.8\% to 89.9\% and 84.1\%, and CLIP-ReID with ViT-B/16 undergoes the largest decline, falling from 91.4\% to 81.7\% and 78.7\%. These results indicate that AdaSP preserves its ranking quality more effectively under visual corruption, while the ViT-B/16 variant is the most sensitive to weather-induced appearance change.

A similar pattern is observed on UAV-VeID-Test, where the setting is more challenging and therefore exposes robustness limitations more clearly. AdaSP drops from 92.7\% mAP to 88.7\% under fog and 76.2\% under rain. CLIP-ReID with ResNet-50 declines from 88.1\% to 83.5\% and 72.4\%, MSINet from 87.7\% to 82.5\% and 71.0\%, and CLIP-ReID with ViT-B/16 from 80.1\% to 70.4\% and 63.3\%. Compared with VRU-Large, the rain-induced degradation is considerably more severe on UAV-VeID-Test for all methods, showing that weather corruption interacts strongly with the inherent difficulty of the evaluation split.

Another important observation is that the performance ordering is largely preserved across weather conditions. AdaSP remains the strongest method, CLIP-ReID with ResNet-50 and MSINet remain competitive, and CLIP-ReID with ViT-B/16 remains the weakest. This consistency suggests that the relative ranking of the compared model families is stable even when the underlying weather condition changes.

The difference between fog and rain is also informative. Fog mainly suppresses contrast and visibility, whereas rain not only reduces clarity but also introduces structured streak noise and additional texture distortion. This dual effect appears to be more damaging to aerial vehicle ReID, particularly when the target vehicles are already small or viewed from oblique angles. As a result, the performance gap between clean and rain conditions is much larger than that between clean and fog conditions across almost all settings.

\subsection{Qualitative Analysis}
The overall performance trends on the two most challenging settings, VRU-Large and UAV-VeID-Test, are summarized in Fig.~\ref{fig:chart}, while representative Rank-5 retrieval examples under adverse weather are shown in Fig.~\ref{fig:Exaple_result}. Together, these figures show that adverse weather affects retrieval both at the dataset level and at the instance level.

As shown in Fig.~\ref{fig:chart}, all methods exhibit a monotonic decline from normal to foggy and then rainy conditions on both VRU-Large and UAV-VeID-Test. AdaSP shows the flattest degradation curve and remains the strongest method under corruption, whereas CLIP-ReID with ViT-B/16 suffers the largest drop. On VRU-Large, AdaSP decreases from 95.9\% to 93.0\% under fog and 88.5\% under rain, while CLIP-ReID with ViT-B/16 drops more sharply from 91.4\% to 81.7\% and 78.7\%. A similar pattern is observed on UAV-VeID-Test, where AdaSP declines from 92.7\% to 88.7\% and 76.2\%, whereas CLIP-ReID with ViT-B/16 falls from 80.1\% to 70.4\% and 63.3\%. The degradation is especially severe on UAV-VeID-Test, where all methods experience a sharper decline than on VRU-Large. This indicates that adverse weather not only reduces retrieval accuracy, but also amplifies the difficulty of already challenging aerial evaluation settings. It also suggests that AdaSP is less sensitive to weather-induced corruption, while CLIP-ReID with ViT-B/16 is the most vulnerable under the current evaluation setting.

The same trend is further confirmed by the retrieval examples in Fig.~\ref{fig:Exaple_result}. Under fog, AdaSP retrieves the correct vehicle at Rank-1, while CLIP-ReID with ResNet-50 and MSINet place it at Rank-2, and CLIP-ReID with ViT-B/16 performs worse. Under rain, the retrieval task becomes more difficult: AdaSP still returns the correct match at Rank-2, whereas CLIP-ReID with ResNet-50 and MSINet place it only at Rank-4, and CLIP-ReID with ViT-B/16 again fails to rank it near the top. 
These examples are consistent with the degradation curves in Fig. \ref{fig:VRU_Big}: steeper mAP drops lead to less stable ranked lists under adverse weather. The larger decline of CLIP-ReID with ViT-B/16 is reflected in its failure to keep the correct match near the top ranks, whereas AdaSP maintains more reliable retrieval under severe weather. This suggests that weather corruption weakens fine-grained local cues and increases confusion with visually similar distractors.

Overall, the degradation curves and qualitative examples show that adverse weather suppresses discriminative cues, shifts rankings toward visually similar distractors, and remains a major bottleneck for UAV-based vehicle re-identification.


\section{Limitations and Conclusion}
This study presents a controlled comparative evaluation of UAV-based vehicle re-identification under simulated clean, foggy, and rainy conditions. Across VRU and UAV-VeID, AdaSP achieves the strongest overall performance, while rain causes the largest degradation. The gap is most pronounced on UAV-VeID-Test, showing that adverse weather further increases the difficulty of challenging aerial retrieval scenarios. These findings confirm that weather corruption remains a major obstacle to reliable UAV-based vehicle retrieval.

This study also has several limitations. First, the weather conditions are synthetically generated rather than collected from matched real-weather UAV trajectories, and may not fully capture real deployment complexity. Second, the comparison covers a limited set of representative methods rather than a broader benchmark. Third, the condition-matched training and evaluation protocol does not assess cross-weather robustness or domain-shift generalization. Future work may explore clean-to-weather and weather-to-weather transfer, additional corruption types and severity levels, and weather-aware training for real UAV deployment.

\section*{Acknowledgment}
This research was supported by the VNUHCM-University of Information Technology's Scientific Research Support Fund.

\bibliographystyle{IEEEtran}
\bibliography{ref}

\end{document}